\documentclass[conference]{IEEEtran}
\IEEEoverridecommandlockouts
\usepackage{cite}
\usepackage{amsmath,amssymb,amsfonts}
\usepackage{algorithmic}
\usepackage{graphicx}
\usepackage{textcomp}
\usepackage{xcolor}
\PassOptionsToPackage{hyphens}{url}\usepackage{hyperref}
\usepackage[ruled]{algorithm2e}
\DontPrintSemicolon
\def\BibTeX{{\rm B\kern-.05em{\sc i\kern-.025em b}\kern-.08em
    T\kern-.1667em\lower.7ex\hbox{E}\kern-.125emX}}
\usepackage{xcolor}
\usepackage[normalem]{ulem}
\begin{document}

\graphicspath{{figures/}}

\title{Graph Neural Networks for Relational Inductive Bias in Vision-based Deep Reinforcement Learning of Robot Control\\}

\author{
\IEEEauthorblockA{Marco Oliva, Soubarna Banik\IEEEauthorrefmark{2}, Josip Josifovski\IEEEauthorrefmark{2} and Alois Knoll}\\
\IEEEauthorblockA{Technical University of Munich, Germany\\ 
Email: \{marco.oliva, soubarna.banik, josip.josifovski\}@tum.de, knoll@in.tum.de}
\thanks{This work has been financially supported by AI4DI project, which
has received funding from the ECSEL Joint Undertaking (JU) under grant
agreement No 826060. The JU receives support from the European Union’s
Horizon 2020 research and innovation programme and Germany, Austria,
Czech Republic, Italy, Latvia, Belgium, Lithuania, France, Greece, Finland,
Norway.\vspace{0.2cm}}
}

\maketitle

\begingroup\renewcommand\thefootnote{\IEEEauthorrefmark{2}}
\footnotetext{These authors contributed equally to this work.}
\endgroup

\begin{abstract}
State-of-the-art reinforcement learning algorithms predominantly learn a policy from either a numerical state vector or images. Both approaches generally do not take structural knowledge of the task into account, which is especially prevalent in robotic applications and can benefit learning if exploited. This work introduces a neural network architecture that combines relational inductive bias and visual feedback to learn an efficient position control policy for robotic manipulation. We derive a graph representation that models the physical structure of the manipulator and combines the robot's internal state with a low-dimensional description of the visual scene generated by an image encoding network. On this basis, a graph neural network trained with reinforcement learning predicts joint velocities to control the robot. We further introduce an asymmetric approach of training the image encoder separately from the policy using supervised learning. Experimental results demonstrate that, for a 2-DoF planar robot in a geometrically simplistic 2D environment, a learned representation of the visual scene can replace access to the explicit coordinates of the reaching target without compromising on the quality and sample efficiency of the policy. We further show the ability of the model to improve sample efficiency for a 6-DoF robot arm in a visually realistic 3D environment.
\end{abstract}

\begin{IEEEkeywords}
graph neural networks, reinforcement learning, robot control, inductive bias, convolutional neural networks
\end{IEEEkeywords}

\section{Introduction}
Most research in the field of reinforcement learning (RL) focuses on \textit{tabula rasa} learning, that is, learning behaviors in environments whose properties, dynamics, and reward landscapes are initially unknown, requiring agents to learn completely from scratch. For many real-work tasks, this requires large amounts of interactions with the environment, making tabula rasa learning neither efficient nor hugely successful in general --- especially when the control of actual hardware is involved. Early approaches to combat sample inefficiency provide an agent with explicitly formulated task-dependent prior knowledge \cite{moreno_using_2004,dixon_incorporating_2000}. Recently, many approaches resort to model-based RL \cite{berkenkamp_safe_2017, kaiser_model-based_2019,zhang_model-based_2020,wang_model-based_2021}.
A different perspective of describing an environment is through entities and their relations. Typically, an agent receives the state of the environment as a flat numerical vector which may include information about objects in the environment but generally does not encode their relations or their importance to the task at hand. This forces the agent to discover such information by thorough exploration during its interactions with the environment. When applying learning algorithms to robotic control applications, however, there exist clear structural relations that remain constant throughout the task and can be explicitly modeled by inducing structural priors into the system. Acknowledging that the robot's structure --- a kinematic chain of actuated joints interconnected by links --- can naturally be modeled as a discrete graph, in which nodes represent joints and edges represent links. This allows the application of graph neural networks \cite{scarselli_graph_2009,battaglia_relational_2018} that directly operate on structural data and exploit the inherent spatial relationships.

\begin{figure*}[t]
\centering
\includegraphics[width=0.99\textwidth]{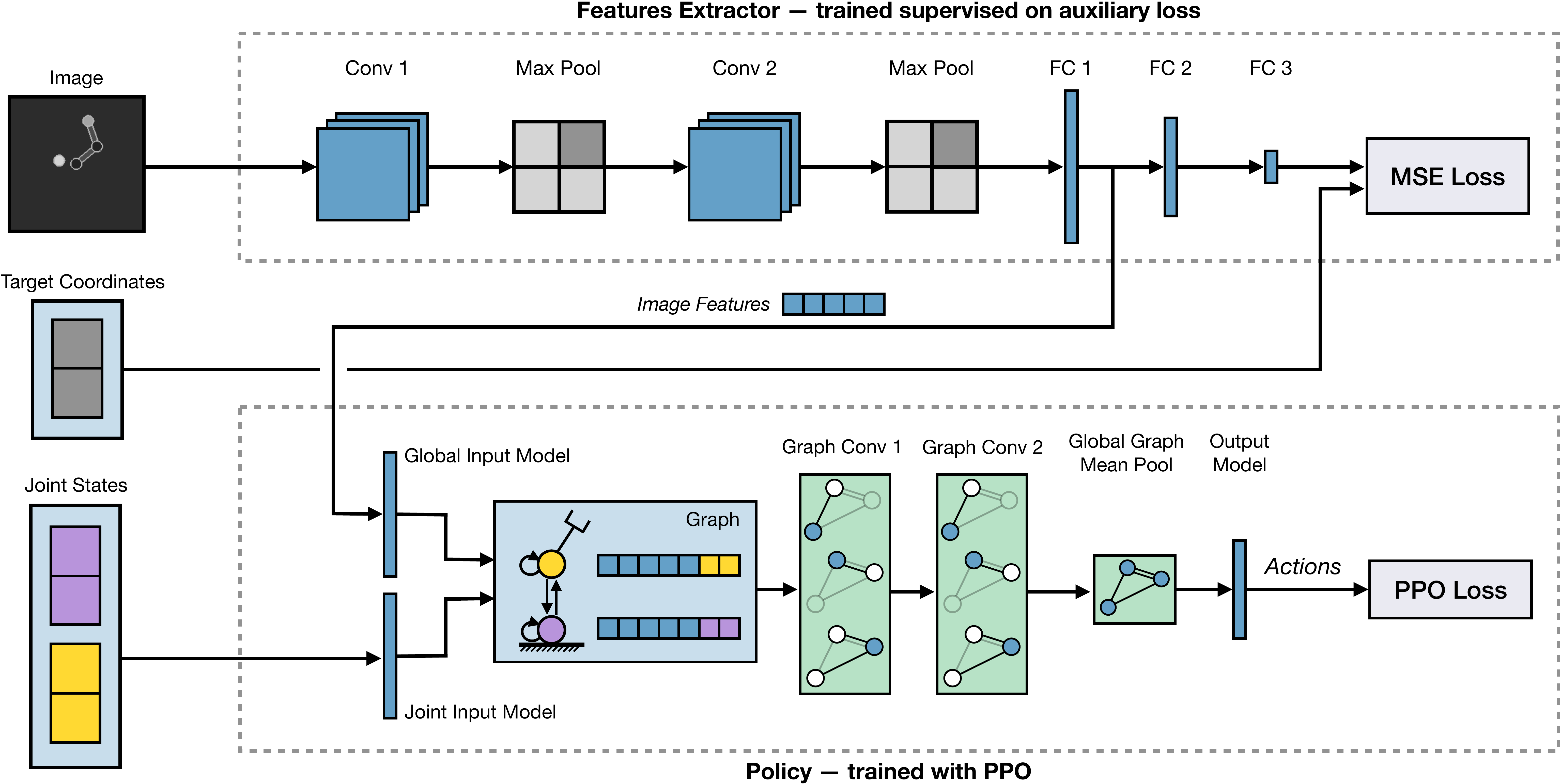}
\caption{An overview of the model architecture: the image encoder processes the image data and extracts an image feature vector which is passed through the global input model to generate a global latent feature vector. It is trained to predict the normalized 2D reaching target coordinates in the image space. The robot's joint states are individually passed through the joint input model to produce a latent feature vector for each joint. From these latent features, a graph representing the current state of the robot and the environment is constructed. The policy graph network learns a control policy mapping from the graph to an $n$-dimensional action vector of angular joint velocity commands.}
\label{fig:cnn-gn-training}
\end{figure*}

In many real-world settings, structural information is not readily available, and visual feedback serves as a mechanism to estimate the environment's state and plan actions. In recent years, many reinforcement learning methods were proposed for image-based settings where a policy is derived from pixel values of the rendered environment and a reward signal, e.g.  \cite{mnih_human-level_2015, zhang_towards_2015, levine_end--end_2016, ebert_visual_2018, yarats_improving_2019}. The growing interest in optimizing control policies from visual feedback is in part motivated by the desire to make the learning methodology more similar to the way that humans learn, i.e. by relying less on exact numerical values and more on vision and combinatorial reasoning. The computational model of graph neural networks that propagates information over relationships introduces this aspect of combinatorial knowledge. Also, in humans, visual feedback alone is insufficient for precise motion. It is well-established that proprioceptive feedback --- the biological pendant to feedback from joint sensors in robots --- plays a crucial role in human motor control and that decreasing the quality or entirely depriving a human of proprioceptive sensory information leads to a significant reduction in motor control precision \cite{bock_method_2007}. Combined with the fact that the vast majority of robots are equipped with highly accurate joint angle and velocity sensors, this motivates extending image-based robotic learning systems to incorporate joint feedback and known kinematic relations.

We identify a blind spot in the reinforcement learning research as, to our knowledge, there exists no method for inducing a relational bias into a vision-based reinforcement learning system. For this reason, we explore and present a graph neural network model that integrates structural knowledge with image observations and apply it to learn an efficient position control policy for robotic manipulation using reinforcement learning.

To this end, we make the following contributions:
\begin{enumerate}
\item We introduce a new architecture that combines an image encoding network with a lightweight graph network to learn an efficient control policy.
\item We show that the relational inductive bias introduced by applying a graph neural network to the graph representation of the robot's state can reduce the sample complexity when training RL agents for robot control without compromising quality.
\item We present a method to iteratively train an image encoding network for image representation learning based on optimizing an auxiliary loss from data generated by the policy.
\end{enumerate}

Supplemental material and code to reproduce the results is available at: \url{https://mrcoliva.github.io/relational-inductive-bias-in-vision-based-rl}.

\section{Related Work}
\label{section:related-work}
With the ever-continuing successes of applying convolutional neural network (CNN) architectures to increasingly challenging problems \cite{yamaguchi_neural_1990,lecun_gradient-based_1998,krizhevsky_imagenet_2012,ciresan_multi-column_2012,szegedy_going_2015}, CNNs have also gained attraction for learning control policies. The works of Mnih et al. \cite{mnih_playing_2013, mnih_human-level_2015} has shown impressive results in learning to play a wide range of Atari games from raw pixel observations and has sparked immense interest in the problem of image-based deep reinforcement learning. Levine et. al also showed in \cite{levine_end--end_2016} that joint end-to-end training of the perception and control systems can improve consistency and generalization, compared to separately trained components, for precise real-world vision-based object placement tasks. With guided policy search methods, they learn a policy mapping from pixels to torques that are applied to a robot. They also induce an algorithmic prior into the system by improving spatial reasoning through a novel spatial feature point transformation.

To circumvent the high cost of real-world data collection in robotics, \cite{James_2019_sim} proposes Randomized-to-Canonical Adaption Networks (RCANs) that learn exclusively on large amounts of labeled data generated in simulation and demonstrate a substantial reduction in required real-world data to produce comparable performance in real-world grasping tasks. To address the difficulty of learning complex behaviors in the partial and high-dimensional image-based observation space, \cite{pinto_asymmetric_2018} introduces an asymmetric actor-critic approach where the actor, i.e. the policy, is limited to visual feedback during inference while the critic is provided with full state observability (images and numerical observations), thus leveraging the full state of the environment during training in a simulator. This leads to improved sim2real performance when deployed on a real robot after training in simulation. Drawing inspiration from this, the model proposed here follows a similar approach by utilizing the simulator to allow the model access to explicit environment states during training time.

Even in simulation, sample inefficiency remains a key challenge in reinforcement learning, especially for model-free approaches. A promising approach to combat this is to exploit a structural bias. Recent strategies to induce bias into learning algorithms are often derived in the form of hierarchical RL, in which agents not only optimize the given reward directly but also utilize \textit{intrinsic motivations}, e.g. to boost exploration in complex environments \cite{kulkarni_hierarchical_2016} or to handle long-term credit assignment problems better \cite{vezhnevets_feudal_2017}. These approaches primarily aim to exploit structure in the task or the environment, rather than directly modeling known relations between entities or even physical relations of the agent itself. A comprehensive review of the importance of relational reasoning to humans and machines alike is given in \cite{battaglia_relational_2018}, where the framework of graph networks as a generalization of graph neural networks \cite{scarselli_graph_2009} is introduced. Especially the introduction of the graph convolutional network (GCN) \cite{bruna_spectral_2013, kipf_semi-supervised_2017} that generalizes the defining characteristics of CNNs --- weight sharing, convolution, and receptive fields --- to structural data where locality is defined as a local neighborhood in the graph domain, could demonstrate the potential of neural network models designed for exploiting relational knowledge. More recently, approaches of combining image features with graph networks were given in \cite{kolotouros_convolutional_2019} applied to human shape recognition with convolutional mesh regression, as well as in the HOPE-Net model \cite{doosti_hope-net_2020} for hand-object pose detection. These works construct the input embeddings of the graph network by concatenating image features extracted from a CNN with some node level features. Our model uses a similar approach to combine image representations with numerical state vectors but instead of offline end-to-end optimization with supervised learning, we train the image encoding network online and separately from a graph network which is trained with reinforcement learning.

An application of graph networks to robotic control is given by the NerveNet model \cite{wang_nervenet_2018}, which directly learns a continuous locomotion control policy for articulated robots in physically realistic simulation environments (\textit{MuJoCo} \cite{todorov_mujoco_2012}) with reinforcement learning. In NerveNet, the kinematic structure of the agent's body is directly mapped into a graph structure, the structural information is propagated through the graph network and finally, joint level actions are predicted by each node. Similar to NerveNet, our approach models the graph as a direct mapping of the agent’s physical structure and trains a graph network with deep reinforcement learning to learn a policy for continuous control of dynamical systems with joint velocity commands. However, unlike NerveNet where the model uses numerical observations, we make use of an image encoding network that produces learned representations of visual observations which we integrate into the graph propagation.

\section{Methods}
\label{section:methods}
We aim to learn an efficient robot control policy from visual observations of the environment and the robot’s readily available internal state information. The model learns to act in an environment by visually detecting relevant properties, without requiring explicit knowledge of their state at inference time. The method can be split into three aspects: (1) extracting task-relevant information from the visual scene; (2) combining the extracted visual information with the internal state information of the robot and processing it in a structured way; (3) learning a control policy for a specific robotic task that maximizes a task-specific reward function. To achieve this, we combine an image encoding network with a policy graph network. The image encoding network learns a transformation of an image of the environment into a low-dimensional image feature vector. The policy graph network learns the control policy from the image feature vector and the robot’s internal states using reinforcement learning. In the remainder of this section, we introduce a robotic reaching task on which the method is evaluated, explain how we induce a relational bias from the robot's known structural information, describe in detail the model architecture, and outline the training procedure.

\subsection{Problem Formulation}
\label{section:task}
\begin{figure}[t]
\centering
\includegraphics[width=0.45\textwidth]{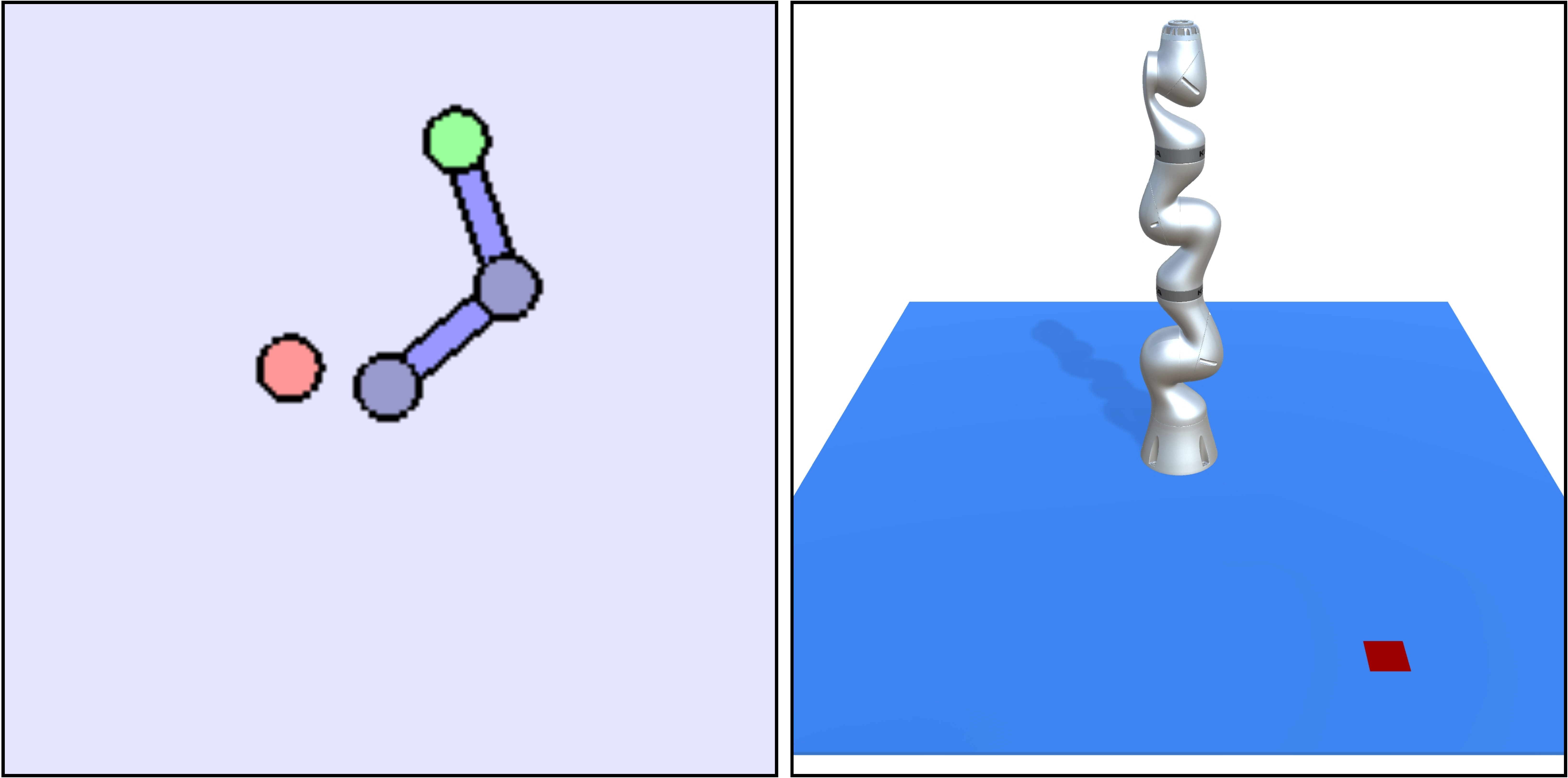}
\caption{Left: the 2-DoF planar robot with the end-effector shown in green in the 2D environment; right: the 7-DoF KUKA LBR iiwa robot in the 3D environment. Note that the viewpoint of the 7-DoF robot in this figure is only for illustrative purposes; the model receives a top-down view instead. The base of both robots is fixed in the center of the environment.}
\label{fig:environments}
\end{figure}
We evaluate our approach on the problem of robotic position control, which entails deriving the inverse kinematics for a robotic manipulator together with a control policy. Starting from an initial robot configuration and a random target position, we train a model that commands joint velocities to move the end-effector of the manipulator to a specified target position. For this, we use two stimulation environments to train and evaluate the model under different levels of environmental complexity (see Fig. \ref{fig:environments}):
\begin{enumerate}
\item A planar $2$-link robot arm in a geometrically simplistic 2D environment \cite{klarmann_investigation_2020}
\item A 7-DoF KUKA LBR iiwa 14 robot in a visually realistic 3D environment \cite{josifovski_continual_2020}. Since the last joint does not contribute to the end-effector position, only the first six joints are considered here and we refer to this robot as 6-DoF.
\end{enumerate}
Both simulation environments implement a Markov Decision Process (MDP) \cite{bellman_markovian_1957} where, at each time step, a state vector consisting of the target position coordinates, the robot's joint sensor states, and an image of the environment is returned, and the agent must perform an action influencing the next state. Thus, we can formulate the task as a reinforcement learning problem:
for the control of a robot with $n$ actuated joints, the model learns a stochastic policy
\begin{equation}
\pi : \mathbf{s}_t \rightarrow \mathbf{a}_t
\end{equation}
mapping from an observation $\mathbf{s}_t = \{ \mathbf{X}_{img}, \mathbf{q} \}$ to an action $\mathbf{a}_t$, where $\mathbf{X}_{img} \in \mathbb{R}^{100 \times 100}$ is a grayscale image of the environment, and $\mathbf{q} \in \mathbb{R}^{2n}$ is the vector containing the internal state of the robot which consists of the current angle $q_i$ and angular velocity $\dot{q_i}$ concatenated for each joint, such that
\begin{equation}
\mathbf{q} = \mathbin {\big \Vert}^n_{i=1} \mathbf{q}_{i} =  [q_1, \dot{q}_1, \dots , q_n, \dot{q}_n] ~.
\end{equation}
In the 3D environment, the agent additionally receives the end-effector coordinates $\mathbf{p} \in \mathbb{R}^3$ in Cartesian space, which can easily be obtained through forward kinematics, such that for each joint
\begin{equation}
\mathbf{q}_{i} = [q_i, \dot{q}_i, p_x, p_y, p_z] ~.
\end{equation}
The policy generates a vector
\begin{equation}
\mathbf{a}_t = [\dot {q}_1^{\prime}, \dots, \dot{q}_n^{\prime}] \in \mathbf{R}^n
\end{equation}
where $\dot {q_i}^{\prime}$ represents the commanded angular velocity for each joint which the simulator converts to actuator torques. 

In each episode, the robot starts from a fixed initial joint configuration and the target position is randomly selected by uniformly sampling a location within the robot’s reachable space. The model should learn a policy that moves the end-effector to the target position as fast as possible. For this reason, we define the reward as the negative distance between the end-effector and the target at each time step, yielding an episode return as the accumulated rewards of all time steps within an episode. This drives the agent to reach the target position with as few time steps as possible. Upon successfully solving the task, the agent receives a high positive reward\footnotemark.

\footnotetext{Based on the specifics of the simulation environments used, the task is solved when the end-effector moves as close as 20 cm to the target position. Upon reaching the target, the agent receives an additional reward bonus of 300 (2D environment) or 10 (3D environment) and the episode terminates, i.e. the agent does not have to hold the position. Episodes also terminate in case the agent fails to reach the target within 300 (2D environment) or 500 (3D environment) time steps. In the 3D environment, the target position is restricted to the ground and represented as a 2D vector on the ground's plane.}

\subsection{Relational Inductive Bias}
\label{section:relational-inductive-bias}
\begin{figure}
    \centering
    \includegraphics[width=0.45\textwidth]{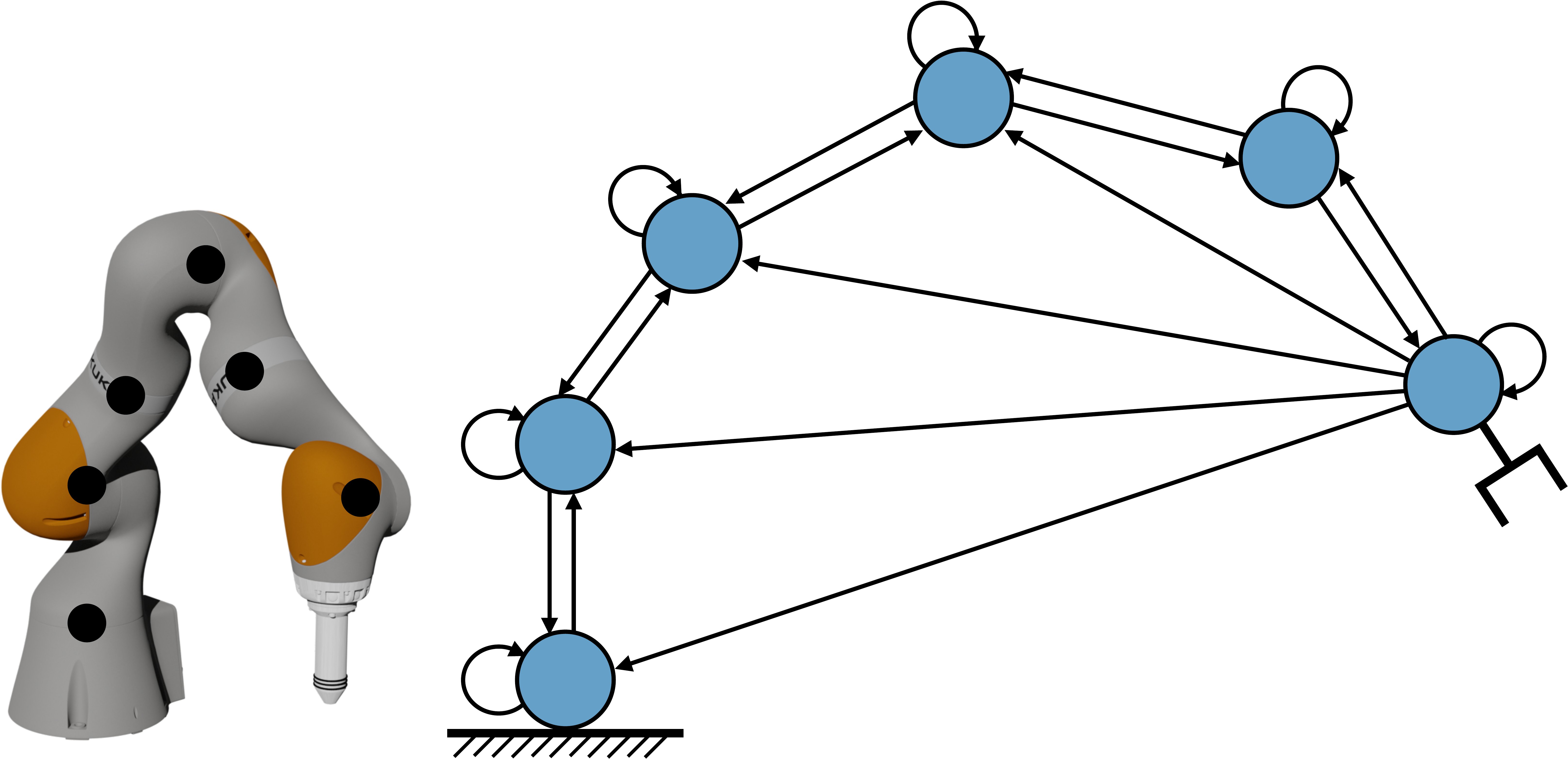}
    \caption{Left: the simulated KUKA LBR iiwa 14 robot with black dots indicating the locations of the six actuated joints contributing to position control. Right: a schematic visualization of the robot's graph representation with six nodes.}
    \label{fig:graph}
\end{figure}

The prior on the structure is implemented as a graph that is constructed analogously to the kinematic structure of the robot arm with each joint being represented by one node and nodes of consecutive joints being connected by edges. Each node has a recurrent edge to itself to receive its state from the aggregation step of the message passing mechanism. Inspired by traditional robot control theory where the Jacobian matrix encodes the effects of joint angle velocities to the end-effector and as a result of qualitative experiments that indicated improved results for robot configurations with a high number of joints, each node additionally receives messages over an incoming edge originating from the node that represents the last controlled joint, i.e. the joint directly connected with the end-effector (see. Fig. \ref{fig:graph}).

\subsection{Model Architecture}
We propose a model consisting of four components (see Fig. \ref{fig:cnn-gn-training}): an image encoding network, a global, as well as a joint input model, and finally a policy graph network. Together, these models implement the following function (for notational simplicity, the subscript $t$ denoting the current time step is omitted in the following derivation): first, the image encoder transforms the input image into an $m$-dimensional global feature vector. Specifically, following the approach in \cite{kolotouros_convolutional_2019}, the image encoder outputs the hidden activations of the first fully connected layer as a low-dimensional encoding of the input image. Inspired by \cite{wang_nervenet_2018}, global and joint-level features are pre-processed by separate input models that are trained jointly with the policy to learn a mapping from the observations into a latent space.

Adhering to the notation of the graph network formalism, we regard the image observation as the global state $\mathbf{u} \in \mathbb{R}^{m}$. Hence, the image encoder $F_{encoder}: \mathbb{R}^{100 \times 100} \rightarrow \mathbb{R}^m$ and the global input model $F_g: \mathbb{R}^m \rightarrow \mathbb{R}^g$ together implement a global update function $\phi^u: \mathbb{R}^{100 \times 100} \rightarrow \mathbb{R}^g$ that transforms $\mathbf{u}$ according to
\begin{equation}
\mathbf{u'} = \phi^u(\mathbf{X}_{img})  =F_{g} (F_{encoder}(\mathbf{X}_{img})) ~.
\end{equation}
Similarly, the joint states $\mathbf{q}_i$ are transformed by the joint input model $F_j: \mathbb{R}^2 \rightarrow \mathbb{R}^j$ given by
\begin{equation}
\mathbf{q}'_i = F_j(\mathbf{q}_i) ~.
\end{equation}
Next, we construct a graph $G = \{ \mathbf{V}, E \}$ that combines global and joint-level latent features by following the approach in \cite{doosti_hope-net_2020}: the input embeddings of each node are obtained by concatenating the global feature vector $\mathbf{u}$ and the respective node-level features $\mathbf{q}^{\prime}_i$, resulting in a node embeddings matrix $\mathbf{V} \in \mathbb{R}^{n \times (g+j)}$ where for each joint $i$:
\begin{equation}
\mathbf{v}_i = \mathbf{u}' \mathbin \Vert \mathbf{q}'_i ~.
\end{equation}
Finally, the policy graph network $\phi^v$  implements a mapping from the graph $G$ to a vector $\mu \in \mathbb{R}^n$:
\begin{equation}
\mu = \phi^v(G) ~.
\end{equation}
We receive a vector of actions $\mathbf{a} \in \mathbb{R}^n$ by sampling from an $n$-dimensional multivariate Gaussian distribution
\begin{equation}
\mathbf{a} \sim \mathcal{N}(\mu, \sigma^2) = \frac{1}{\sigma\sqrt{2 \pi}} e^{-\frac{1}{2}\big(\frac{\mathbf{a} - \mu}{\sigma}\big)^2}
\end{equation}
where $\mu$ is the vector of the means and $\sigma \in \mathbb{R}$ is a learned parameter giving the standard deviation. We now give additional details about the individual models.

\subsubsection{Input Models}
Both the global input model and the joint input model are implemented as a single fully-connected layer with a $tanh$ activation to produce the input features of the policy and value network. The global input model $F_g: \mathbb{R}^{128} \rightarrow \mathbb{R}^{128}$ transforms the image feature vector produced by the image encoder to a 128-dimensional global feature vector.

Likewise, the joint input model $F_j: \mathbb{R}^{2} \rightarrow \mathbb{R}^{32}$ learns a mapping from the 2D joint states (angular positions and velocities) to a 32-dimensional joint feature vector. When training the model on the 3D environment, the input to the joint input model additionally contains the robot's end-effector coordinates in Cartesian space, yielding $F_j: \mathbb{R}^{5} \rightarrow \mathbb{R}^{32}$.

\subsubsection{Image Encoder}
The image encoder is implemented as a CNN that receives an image of the environment from an overhead viewpoint and learns to detect the reaching target and predict the 2D coordinates in the environment. The observed RGB image is pre-processed into a normalized grayscale image. Due to the simplicity of the target detection task, we use a simple network structure with two convolution blocks, each consisting of a convolution with a kernel of size $5 \times 5$ and stride $1$, ReLU activations, and a max-pooling layer of size $2 \times 2$ and stride 2. The number of filters in the convolution layers is 6 and 16 respectively. This results in a flattened hidden representation of size $7744$ which is passed through a fully connected layer (FC 1) with ReLU activations and $m$ output units, where $m$ is the dimension of the image feature vector. At inference time, the resulting image features are passed to the policy. In contrast, during training the image features are further propagated through a prediction head, consisting of a fully connected layer with 84 units and ReLU activations, and an output layer with 2 units and $tanh$ activations that outputs the estimates of the target's $x$ and $y$ coordinates in the normalized world coordinate space (see Fig. \ref{fig:cnn-gn-training}). Hence, we do not use this estimate of the target coordinates to replace the policy's access to the true coordinates, but instead provide a latent encoding of the image to avoid restricting the application of the model to tasks where only a specific feature set is relevant. Qualitative experiments showed best results when using the output of FC 1 as the image encoding, compared to other hidden representations in the network.

\subsubsection{Policy Graph Network}
The policy graph network is a simple spectral-based graph convolutional network \cite{kipf_semi-supervised_2017}, performing the node update
\begin{equation}
f(\mathbf{V}^{(l)}, \mathbf{A}) = \sigma \bigg ( \tilde{\mathbf{D}}^{-\frac{1}{2}} \tilde{\mathbf{A}} \tilde{\mathbf{D}}^{-\frac{1}{2}} \mathbf{V}^{(l)} \mathbf{W}^{(l)} \bigg ) ~,
\end{equation}where $\mathbf{V} \in \mathbb{R}^{N \times D}$ is a matrix constructed by stacking the embeddings $\mathbf{v}_i \in \mathbb{R}^D$ of all $N$ nodes, $\tilde{\mathbf{A}} = \tilde{\mathbf{D}}^{-\frac{1}{2}} \mathbf{A} \tilde{\mathbf{D}}^{-\frac{1}{2}}$ where $\tilde{\mathbf{D}} \in \mathbb{R}^{N \times N}$ is the diagonal node degree matrix and $\mathbf{A} \in \mathbb{R}^{N \times N}$ is the adjacency matrix, and $\mathbf{W}^{(l)}$ is the weight matrix of layer $l$. 

The network receives an input graph with a node embeddings matrix $\mathbf{V}^{(0)} \in \mathbb{R}^{n \times 160}$ where $n$ is the number of nodes, i.e. each node has an initial hidden state of size 160 which contains the concatenated outputs of the global input model (128-dimensional) and the joint input model (32-dimensional). Using two graph convolutional layers with a hidden dimension $h = 256$ and ReLU activations, the network produces an updated embeddings matrix $\mathbf{V}^{\prime} \in \mathbb{R}^{n \times h}$, followed by a global graph mean pooling operation over all $n$ nodes that reduces the embeddings to a global graph state $\mathbf{g} \in \mathbb{R}^h$ with
\begin{equation}
\mathbf{g} = \frac{1}{n} \sum^n_{i} \mathbf{v}_i^{\prime} ~.
\end{equation}
Ultimately, an output model $F_{out}: \mathbb{R}^h \rightarrow \mathbb{R}^n$ consisting of a single linear layer maps the global graph state down into the action space to an output vector $\mu \in \mathbb{R}^n$ that represents the means of the multivariate Gaussian distribution from which actions are sampled:
\begin{equation}
\mathbf{\mu} = F_{out}(\mathbf{g})
\end{equation}
\subsubsection{Value Function}
The state-value function $f_v$ is implemented as a three-layer MLP. It receives as input the latent global state concatenated with all latent joint states, i.e. a vector in $\mathbb{R}^{128 + 32n}$, which passes through two hidden layers with 256 units and ReLU activations, and a linear output layer that outputs a scalar value. As a result, the state-value is computed according to
\begin{equation}
v = f_v(\mathbf{u}^{\prime} \mathbin \Vert \mathbf{q}') ~.
\end{equation}

\begin{algorithm}
Initialize image encoder parameters $\theta_{encoder}$\;
Initialize policy parameters $\theta_{policy}$\;
\;
\While{maximum time steps not reached}{
 $\{ \mathbf{A}, \mathbf{O}, \mathbf{r} \} =$  \texttt{collectRollout()} \tcp*{(actions, observ., rewards)}\;
 
\tcp{Optimize the policy network}
\For{each policy epoch}{
\For{each batch}{
$\theta_{policy} \leftarrow$ \texttt{ppoStep}$(\mathbf{A},\mathbf{O},\mathbf{r})$\;
}}

\tcp{Optimize the image encoder}
\For{each encoder epoch}{
\For{each batch}{
 $\mathbf{Y} \leftarrow$ \texttt{extractTargetCoord}$(\mathbf{O})$\;
 $\mathbf{I} \leftarrow$ \texttt{extractImageData}$(\mathbf{O})$\;
 $\hat{\mathbf{Y}} \leftarrow$ \texttt{imageEncoderForward}($\mathbf{I}$)\;
 $\theta_{encoder} \leftarrow$ \texttt{optimizationStep}$(\hat{\mathbf{Y}},\mathbf{Y})$\;
}}
 }
\caption{Training with auxiliary loss optimization}
\label{alg:cnn-gn-aux-opt}
\end{algorithm}

\subsection{Training}
\subsubsection{Policy}
The policy, comprised of the input models and the policy network, is trained with PPO \cite{schulman_proximal_2017}. We use the PyTorch \cite{paszke_pytorch_2019} implementation provided by \textit{Stable Baselines 3} \cite{raffin_stable_2019}. PPO is a policy gradient method where the agent alternates between sampling data from the environment, i.e. collecting a rollout and performing an optimization step to minimize the PPO loss which is given by
\begin{equation}
\begin{split}
L(\mathbf{\theta}) = \hat{\mathbb{E}} \bigg [ \min \bigg(r_t(\theta) ~\hat A_t, \mathrm{clip}(r_t(\theta), 1 - \epsilon, 1 + \epsilon) ~\hat A_t \bigg) \\ 
- c_1 L_t^{VF}(\mathbf{\theta}) + c_2 S[\pi_{\theta}](s_t) \bigg ] ~,
\end{split}
\end{equation}
where $r_t(\theta)$ denotes the probability ratio between the new and the old policy, $\hat A_t$ denotes the advantage function, $\epsilon$ defines the size of the clipping region, $c1, c2$ are weighting coefficients, $L_t^{VF} = (V_{\theta}(s_t) - V_t^{\mathrm{target}})^2$ is a squared-error loss for the value function, and $S$ is an entropy bonus that encourages sufficient exploration \cite{schulman_proximal_2017}. The hyperparameters of PPO are defined as $\epsilon = 0.2$, $c_1 = 0.5$, and $c_2 = 0$.
We follow the recommendations in \textit{OpenAI} \textit{Stable Baselines} \cite{dhariwal_openai_2017} for weight initialization in PPO agents and initialize the network weights as random (semi) orthogonal tensors \cite{saxe_exact_2014}. All weight matrices are element-wise multiplied with a gain value of $\sqrt{2}$, with two exceptions: the output layer of the value network uses a gain of $1.0$, and the output layer of the policy network uses a gain of $0.01$. For other hyperparameters, we performed an automated search on the 2D environment and picked the parameters that performed best across all architectures. Using 20 environments in parallel, the agent performs 256 time steps on each environment, yielding rollouts consisting of $5120$ elements. Optimization is performed by partitioning the rollout buffer into 32 mini-batches of size 160 and training each rollout over 4 epochs with an initial learning rate $\alpha_0 = 0.00025$. The learning rate follows the decay rule $\alpha_t =  \alpha_0 \min \left ( 1, \left( \frac{5}{4} - \frac{t}{T} \right) \right)$ where $t$ is the current time step and $T$ is the total number of time steps performed during training. From this rule, the learning rate remains initially constant until $t = \frac{T}{5}$ and then linearly decays until $t=T$, such that $\alpha_T = \frac{1}{5} \alpha_0$. After each optimization step, the rollout buffer is deleted and a new rollout is collected for the next optimization step.

\subsubsection{Image Encoder}
\label{section:encoder_training}
Rather than training the image encoder jointly with the policy in an end-to-end reinforcement learning process, it is trained, similarly to the policy, after each rollout collection period, but by optimizing an auxiliary loss function based on the rollout data in a separate supervised learning procedure (see Alg. \ref{alg:cnn-gn-aux-opt}). The training dataset for the image model is constructed by extracting from the observations in the rollout buffer the image observations as well as the corresponding 2D coordinates of the reaching target to get input-output pairs for each time step of the rollout. The encoder is then optimized using Adam to predict the reaching target coordinates, i.e. to minimize the mean-squared-error between the predicted coordinates and the ground-truth.

\section{Experiments}
\label{section:experiments}
We shall refer to our model as the CNN-GN model. To demonstrate and give context to the effectiveness of our architecture, we compare the CNN-GN model performance with the following baseline models:

\begin{itemize}
    \item \textbf{MLP} A two-layer MLP model that receives the full flat state vector, including target coordinates, without images.
    \item \textbf{GN} A two-layer GN that receives the full state but converts it into a graph representation before processing it. This baseline is insightful for analyzing the isolated effect of a GN processing structural information over the standard MLP, as well as the impact of replacing the target coordinates with images in the CNN-GN model.
    \item \textbf{CNN-MLP} A model that is identical to the CNN-GN model, receiving the same state with images and the robot's state, but using an MLP policy network instead of a GN. This allows analyzing the effect of the relational inductive bias of the GN when learning also with images.
    \item \textbf{CNN-MLP-IMG} A model that receives only images, i.e. applies a two-layer MLP to the global input model's output, which allows inferring the benefit of access to the agent's internal state over a vision-only approach.
\end{itemize}
All models are evaluated by running experiments on both simulated continuous control tasks (see Section \ref{section:task}). The results are obtained by averaging the rewards of running each experiment five times with different random seeds. The models are trained over one million time steps and periodically evaluated on  separate evaluation environments on which actions are selected deterministically as the mode of the distribution generated by the policy network. Table 1 reports a summary of the results of all models in both tasks.

\begin{table*}
\begin{center}
\begin{tabular}{| c || c | c | c || c | c |}
\hline
Input & \multicolumn{2}{c|}{images + internal states} & images & \multicolumn{2}{c|}{full numeric state} \\
\hline
{Model} & CNN-GN & CNN-MLP & CNN-MLP-IMG & GN & MLP \\
\hline\hline
2-DoF task & $250 \pm 22.7$ & $\mathbf{251 \pm 21.1}$ & $175 \pm 42.1$ & $262 \pm 13.4$ & $\mathbf{272 \pm 9.3}$ \\
\hline
6-DoF  task & $\mathbf{-94 \pm 22.9}$ & $-102 \pm 26.3$ & $-408 \pm 52.2$ & $\mathbf{-61 \pm 15.9}$ & $-63 \pm 14.3$
\\
\hline
\end{tabular}
\end{center}
\caption{The average rewards and standard deviations that each model converged to in both environments,  reported over five runs per model by averaging rewards and standard deviations across runs, but over the last last 100k time steps for the latter. The best scores are highlighted in bold. Note that rewards are not comparable between environments, i.e. in the 2D environment, a reward of $270$ is near optimal, while in the 3D environment, a comparably well-performing policy achieves a reward of $-50$.}
\label{table:results}
\end{table*}

\subsection{2D position control of a planar 2-DoF manipulator}
\begin{figure}[t]
\centering
\includegraphics[width=0.44\textwidth]{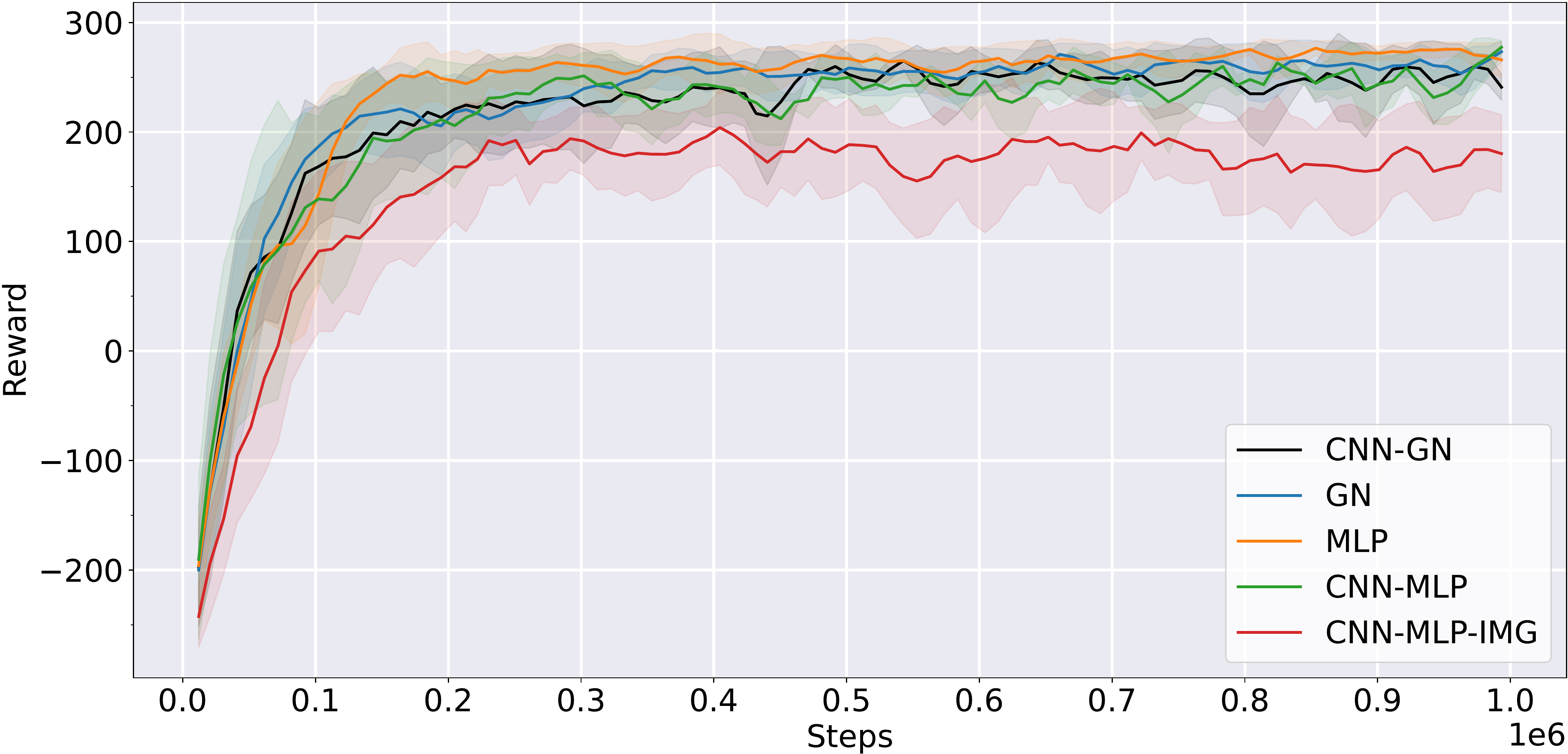}
\caption{Visualization of the average rewards achieved on the evaluation environments over time of the proposed CNN-GN model (black) and all baselines models on the 2D position control task. The shaded region around the line plots visualizes the standard deviation within the data points generated by five runs per model.}
\label{fig:results_2link}
\end{figure}

Fig. \ref{fig:results_2link} visualizes the achieved rewards of the policies over one million time steps on the 2D environment. The CNN-GN model learns an efficient policy and converges to a mean episode reward of more than 250 (see Table \ref{table:results}), which amounts to a near-optimal policy on the environment. The model can match the performance of the GN baseline model, both in quality and sample complexity\footnotemark\footnotetext{The GN and MLP models serve as an upper performance bound since they receive the full environment state without the need to infer it from images.}. There is no significant difference between the CNN-GN and CNN-MLP models (see Table \ref{table:results}), i.e. no apparent benefit of using a graph network for the agent's policy. The MLP baseline model reaches marginally higher rewards than other baselines. On the other hand, the CNN-MLP-IMG model as the only model without access to the robot's internal state is unable to learn a policy of the same quality than the CNN-GN model and other baselines.

\subsection{3D position control of a 6-DoF manipulator}
\begin{figure}[t]
\centering
\includegraphics[width=0.44\textwidth]{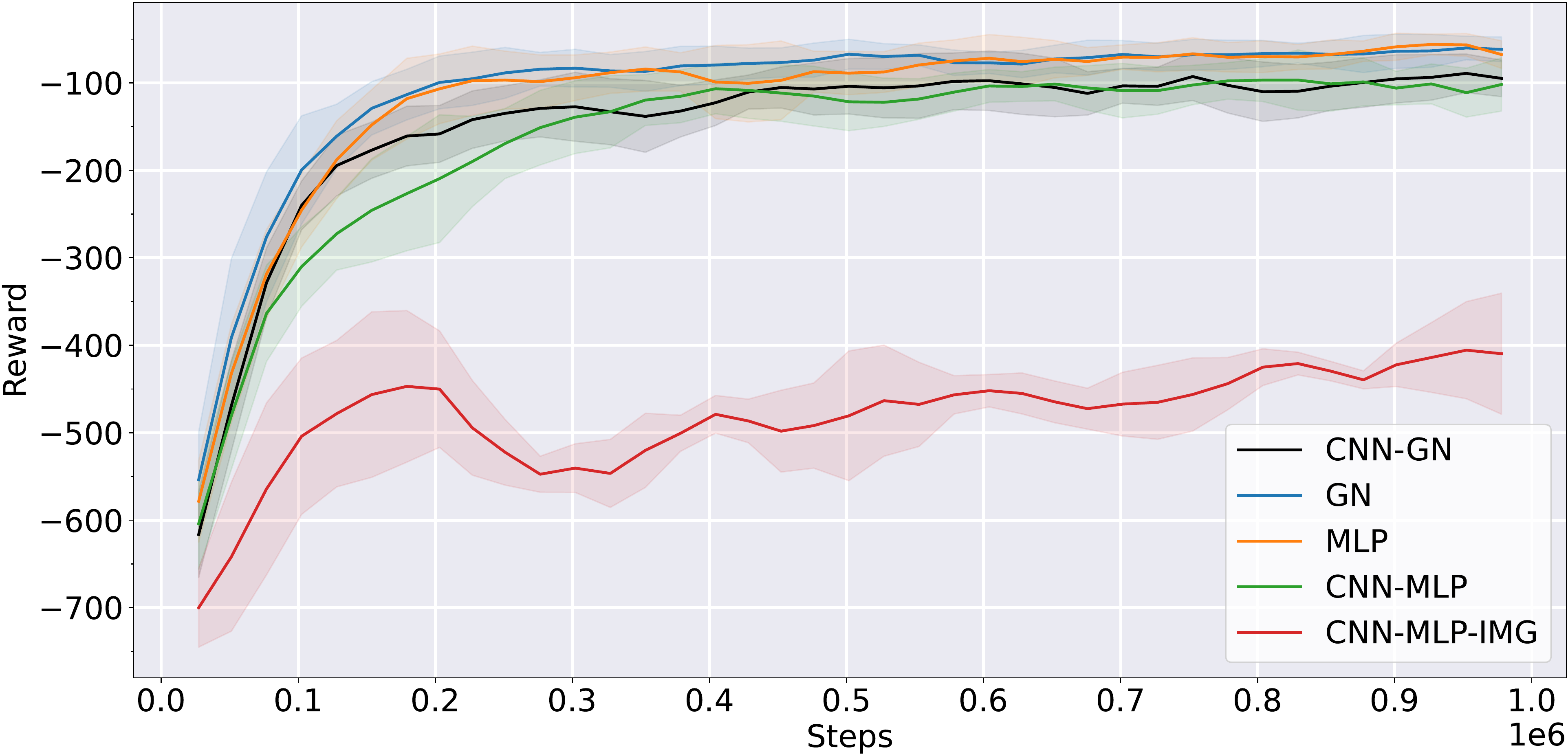}
\caption{Reward plots of the proposed CNN-GN model (black) and all baselines models on the 3D reaching task with the 6-DoF manipulator. The shaded region around the line plots visualizes the standard deviation within the data points generated by five runs per model.}
\label{fig:results_6link}
\end{figure}
On the 3D position control task, the model is also capable of learning an efficient control policy for a 6-DoF robot manipulator, although, not fully reaching the performance of the GN and MLP baseline models that have direct numerical access to the full environment state including the target position (see Fig. \ref{fig:results_6link}). While they arrive at a near-optimal policy with average rewards of around -60 after 400-500k time steps, the CNN-GN model converges to $-94$ after one million time steps, which is marginally better than the CNN-MLP baseline converging to $-102$. The CNN-MLP-IMG baseline model is incapable of solving the task solely relying on image observations. Hence, after training for a million time steps, the CNN-GN model reaches the target on average in roughly 85\% of episodes. Although the same is true for the CNN-MLP baseline, the policy learned by the CNN-GN model learns faster initially, before the CNN-MLP policy effectively reaches performance parity after around 325k time steps.

\subsection{Discussion}
The results demonstrate that the proposed model is capable of learning successful control policies in both environments. For the 2-link planar robot, the graph representation with only two bidirectionally connected nodes does not significantly increase the structural information given to the agent, which is why a notable improvement over an MLP policy is not to be expected. And indeed, the MLP model performs generally best on the 2D environment, which is in line with the findings of \cite{wang_nervenet_2018} for a comparable reaching task. The performance of the CNN-GN model is equivalent to the CNN-MLP baseline model within the noise expected from the inherent randomness of the process. These results indicate no apparent benefit of the relational bias induced by the graph representation in the 2-link case, both for the numerical case and when learning with images. On the other hand, the fact that the CNN-MLP-IMG agent learns a partially successful policy only from the image encoding suggests that the latent image features produced by the image encoder not only contain information about the location of the reaching target, but contribute valuable statistics about the general state of the environment, including the robot. Nonetheless, the policy learned from images alone is still evidently worse than the other models, which strongly confirms the benefit of providing the control algorithm with the robot's internal state over the exclusive usage of vision.

In contrast, in the 3D environment the CNN-GN model learns visibly faster than the CNN-MLP baseline early on during training. The marginal advantage of the MLP baseline model on the 2-DoF robot is not present on the 6-DoF robot, indicating that the relational bias induced by the graph representation improves the sample efficiency of the learning algorithm when applied to larger graphs. This likely stems from the more informative relational bias encoded in the richer graph representation of the 6-DoF robot compared to the 2-DoF robot. Nevertheless, the results don't show a significant difference in the quality of the converged policy after training for one million time steps, which concludes that the graph network can learn a good policy faster, but does not, in general, converge to a better policy.

Unlike in the 2D environment, the CNN-GN model and the CNN-MLP baseline achieve moderately lower rewards compared to their GN and MLP counterparts, due to the more challenging vision task in 3D. On the other hand, the CNN-MLP-IMG model is incapable of learning a successful policy.  This suggests that the top-view of the environment is sufficient to derive a planning strategy, i.e. robustly detecting the target location, when the robot's state is available but, due to the potential occlusions of large parts of the robot and potentially the target, it is not enough to fully decode the joint states which in turn appears to limit the quality of the resulting control policy.

\subsection{Ablation studies}
\begin{figure}[t]
\centering
\includegraphics[width=0.42\textwidth]{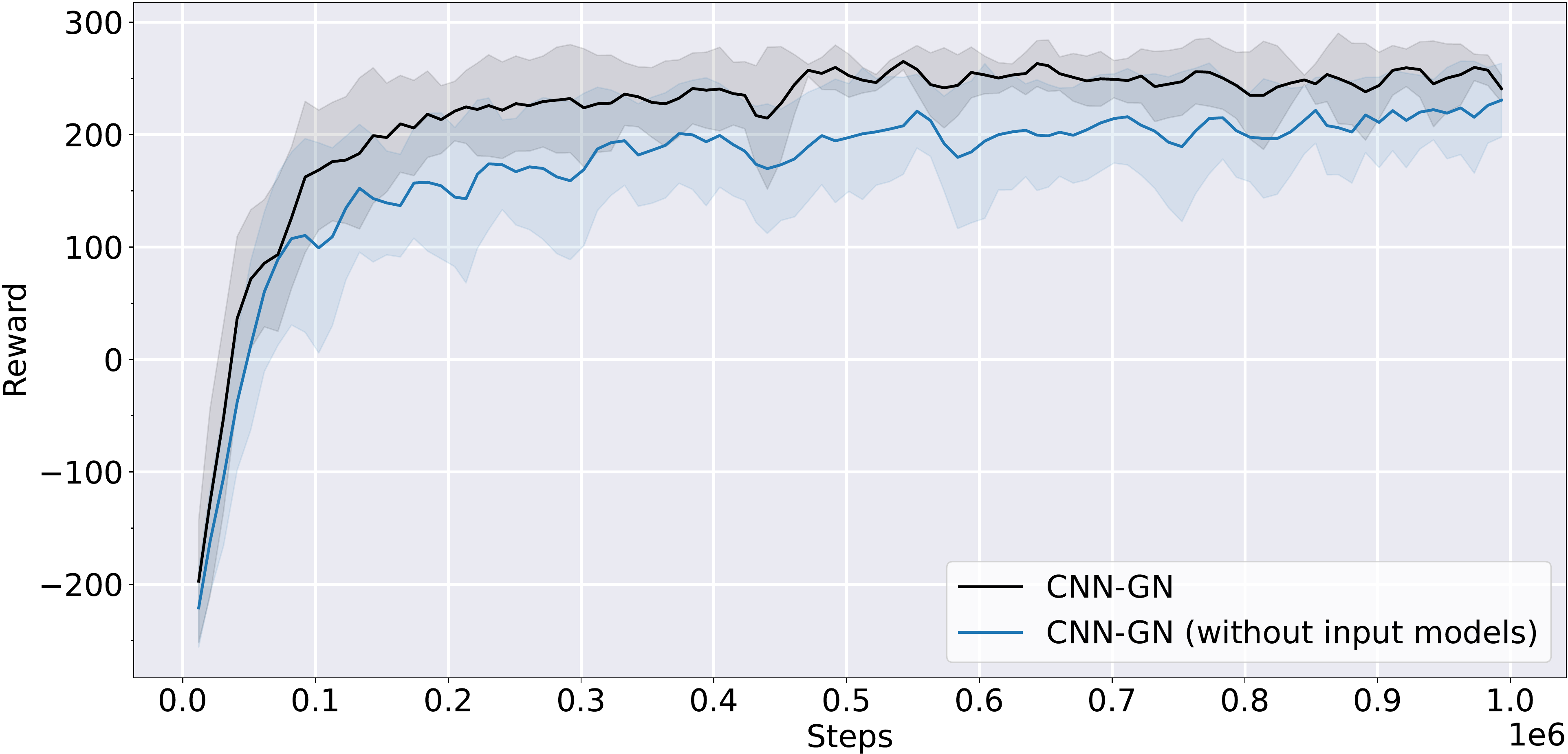}
\caption{Reward plots for the 2D task of the proposed CNN-GN model which uses input models (black) and a variant without input models (blue).}
\label{fig:input_models_effect}
\end{figure}

In addition to the experimental results outlined above, we conducted several ablation studies of variations to the proposed architecture to justify the design choices made. Most notably, we trained a model that omits the pre-processing of observations by global and joint input models but is otherwise identical to the CNN-GN model. The results showed that the use of input models leads to consistently higher rewards throughout the training process (see Fig. \ref{fig:input_models_effect}).

\begin{figure}[t]
\centering
\includegraphics[width=0.42\textwidth]{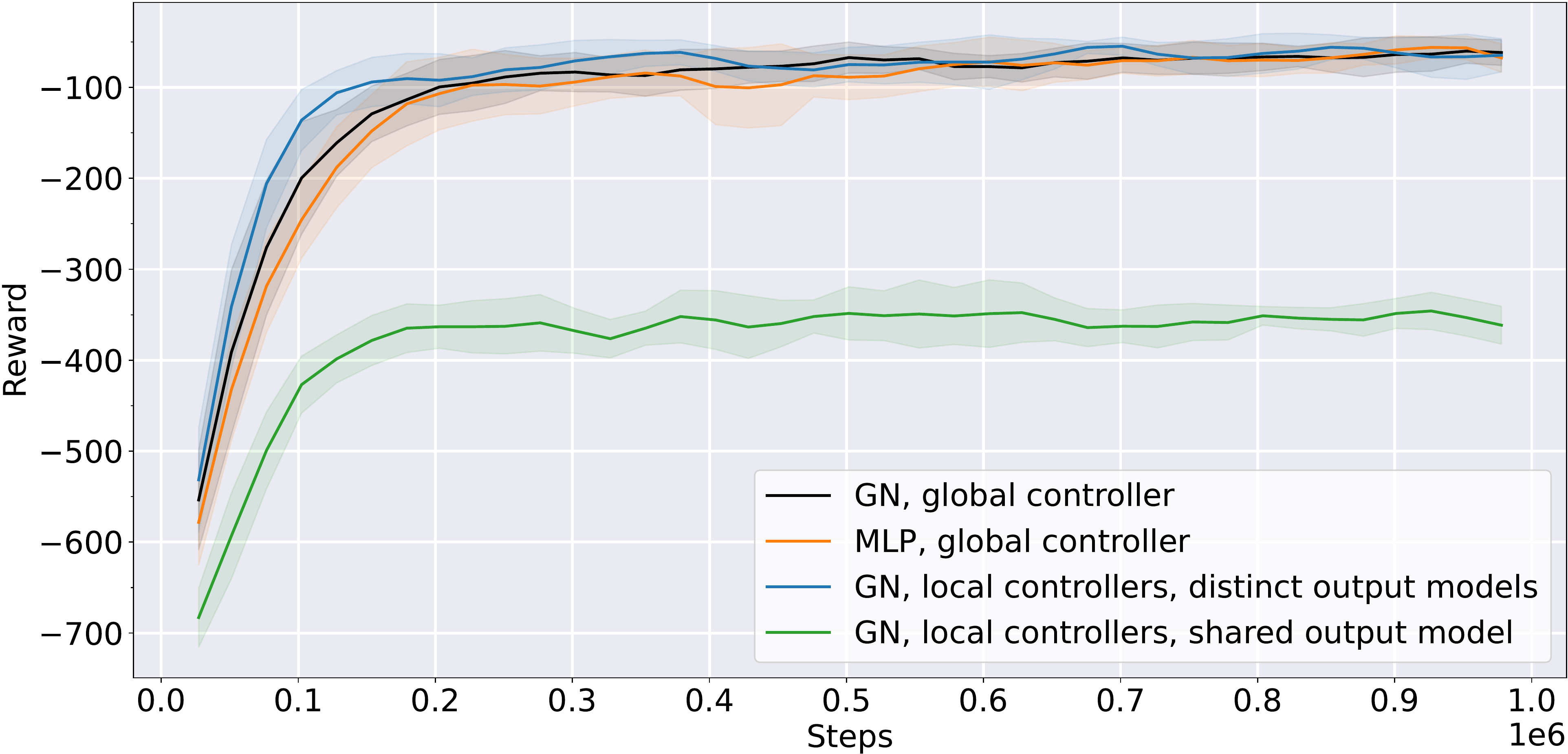}
\caption{Reward plots for the 3D task of the GN and MLP baseline models (black, orange) and two variants of the GN model where the joint velocities are computed only from the embeddings of the corresponding node.}
\label{fig:GN_node_level_actions}
\end{figure}

Furthermore, we evaluated two alternative variants of the GN baseline on the 3D task where the models learn local, decentralized controllers at each node, in contrast to the global controller realized by the global graph pooling operation. One approach is that actions are computed by individually feeding the output embeddings of each graph node into a shared output model, effectively producing actions directly from each node's state. However, agents generating node-level actions this way performed significantly worse across experiments in both the 2D and 3D tasks. In contrast, in the second approach, each node learns a local controller by using a dedicated output model which was able to learn a good policy faster and converge earlier than the GN and MLP models with the global controller, although at a considerable computational cost that hinders scaling the method to systems with many degrees of freedom. Fig. \ref{fig:GN_node_level_actions} shows the respective learning curves.

\section{Conclusion}
\label{section:conclusion}
We explored a blind spot in the reinforcement learning research: the combination of inducing a relational bias of the known structure underlying the task with image-based observations to learn a continuous robot control policy. To this end, we presented a model architecture that combines visual feedback with the agent's state encoded in a graph representation of its physical structure. While the relational bias induced by the graph does not improve the sample efficiency of a control policy for a simple 2-DoF robot, we could demonstrate that it does indeed enable faster learning on a larger system in the form of a 6-DoF manipulator, given that the graph representation is rich enough to provide considerable structural information compared to a flat observation vector. Further, we conclude that the image features produced by the image encoder can replace explicit access to the target position coordinates without compromising on quality or sample efficiency in case the image theoretically fully describes the environment, as is the case in the 2D environment. When this assumption is violated, for example, due to occlusions or viewpoints that only enable partial observability, the proposed system could still learn a very successful policy, yet not fully matching the results of an agent that has access to the full numerical state.

An obvious next step is to apply the proposed method to a real robot to validate sim2real transfer, as well as to robotic tasks with higher structural complexity, like manipulating objects, to validate the conclusion that the method's usefulness increases with increasing structural system complexity. The asymmetric training where the image encoder has access to the reaching target coordinates during training time currently limits the application to tasks where such ground truth data is readily available. While this is easy for simulated environments, a direct transfer to real-world tasks is unlikely to perform well without a considerable amount of manual labeling. For this reason, applying unsupervised representation learning strategies, such as using a variational auto-encoder for image feature extraction, is a promising direction for future work to better bridge the reality gap.

\section{Acknowledgement}
The authors would like to thank Mohammadhossein Malmir and Noah Klarmann for their contribution in the development of the simulation environments and Stefan Böhm for reviewing the content.
\bibliographystyle{IEEEtran}
\bibliography{references.bib}

\end{document}